\documentclass{article}
\usepackage{spconf,amsmath,graphicx}
\usepackage{enumitem}
\usepackage{hyperref}
\usepackage{color,array}

\title{What Happens Without Background?  
Constructing Foreground-Only Data for Fine-Grained Tasks}
\name{Yuetian Wang, Wenjin Hou,  Qinmu Peng and Xinge You$^\dag$
\thanks{
\dag  Corresponding author
}
}
\address{School of Electronic Information and Communications, Huazhong University of Science and Technology}

\begin{document}
%
\maketitle
\begin{abstract}
Fine-grained recognition, a pivotal task in visual signal processing, aims to distinguish between similar subclasses based on discriminative information present in samples. 
However, prevailing methods often erroneously focus on background areas, neglecting the capture of genuinely effective discriminative information from the subject, thus impeding practical application. 
To facilitate research into the impact of background noise on models and enhance their ability to concentrate on the subject's discriminative features, we propose an engineered pipeline that leverages the capabilities of SAM and Detic to create fine-grained datasets with only foreground subjects, devoid of background. 
Extensive cross-experiments validate this approach as a preprocessing step prior to training, enhancing algorithmic performance and holding potential for further modal expansion of the data.
The code and data will be available at:
\end{abstract}
\begin{keywords}
Computer Vision, Deep Learning, Fine-grained visual categorization, Segment Anything Model
\end{keywords}
\section{Introduction}
Fine-grained recognition stands as a cornerstone in the field of computer vision\cite{xu2023mmcosine}, tasked with the discernment of subtle differences between closely related subclasses within a category. This capability is paramount for applications ranging from biodiversity conservation to medical diagnostics\cite{he2022transfg}, where the accurate identification of species or conditions can have profound implications.

\begin{figure}[tbp]
\includegraphics[width=0.5\textwidth]{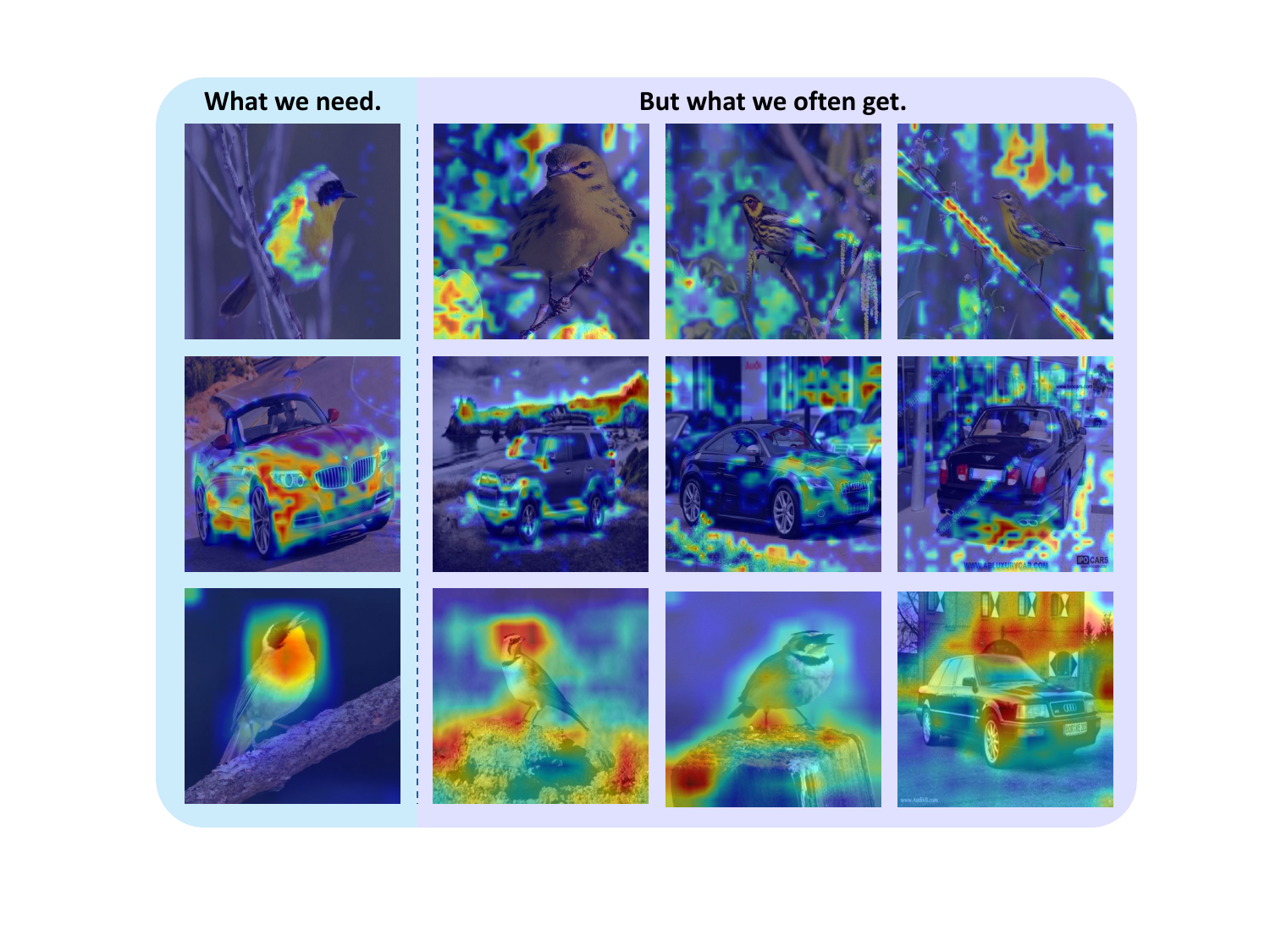}
\caption{\textbf{Grad-CAM Visualization} of Common Backbones in Fine-Grained Classification: the first two rows for ViT and the last row for ResNet.} 
\label{fig_cam}
\end{figure}

Despite certain progress in related research, existing methods are often hampered by an erroneous focus on background elements, which detracts from the extraction of key discriminative cues from the subject itself\cite{zha2023boosting}. 
As illustrated by the Grad-CAM\cite{selvaraju2017grad} visualisation in Figure \ref{fig_cam}, there is an undesired focus on non-subject content, regardless of whether CNN-based or Transformer-based backbones are used. This misfocus limits model accuracy and real-world applicability, as the discriminative information on the subject is the truly effective element.

In light of these issues,  this study introduces an innovative engineered pipeline to facilitate further research into the impact of background noise on model performance and enhance the model's ability to focus on the subject's genuine discriminative features. By leveraging the capabilities of SAM\cite{kirillov2023segment} and Detic\cite{zhou2022detecting}, we have constructed a dataset that isolates the foreground subjects, thereby providing a more refined and accurate environment for model training and evaluation. Further, we combine a variety of different commonly used Backbone explorations done on the new data obtained.

The contributions of this paper can be summarized as follows:
\begin{itemize}[leftmargin=15pt,topsep=3pt]
    \item Firstly, we demonstrate through visualization that existing fine-grained models have a high probability of being influenced by the background, incorrectly focusing on non-subject regions at the expense of learning truly valid discriminative information, thereby hindering practical applicability.
    \item To address this issue, we propose an engineered pipeline utilizing Visual Foundation Models to generate foreground-only data, facilitating research into the impact of background noise and improving the model's capacity to capture discriminative subject features.
    \item Furthermore, this pipeline can be used as a general data pre-processing procedure in fine-grained tasks, producing more densely distributed and higher quality training data. Comprehensive cross-validation experiments confirm the performance boost from the preprocessing process.
    \item Additionally, we propose that the generated foreground data can also serve as the foundation for expanding the corresponding datasets into more modalities.
\end{itemize}

\section{Data Construction}
\subsection{Pipeline}
Our goal is to generate foreground data efficiently through a streamlined and automated pipeline that minimizes manual effort and ensures high-quality results. The pipeline is designed to be both highly automated and versatile, enabling dataset creation across different domains without specific annotations, training, or fine-tuning.

\begin{figure}[tbp]
\includegraphics[width=0.48\textwidth]{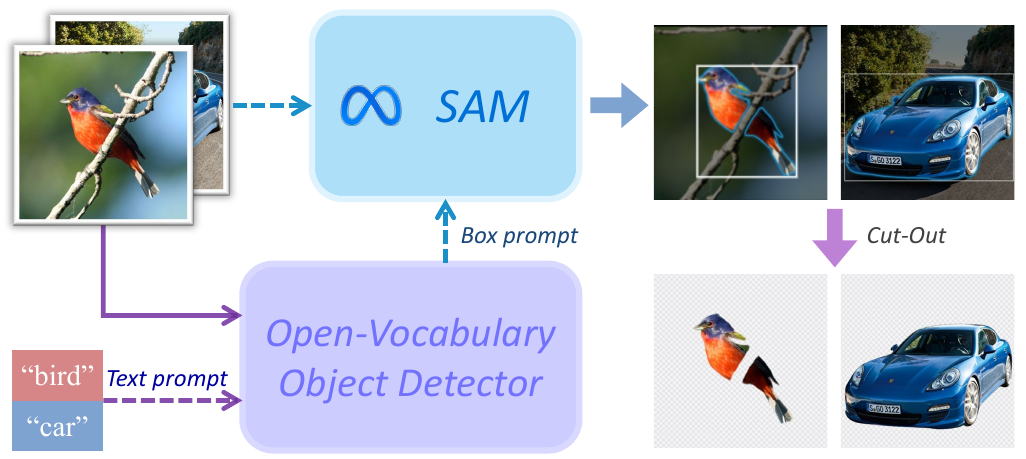}
\centering
\caption{Proposed Pipeline for Generating Foreground Images.} 
\label{fig_pip}
\end{figure}

We utilize the Segment Anything Model (SAM) for meticulous object segmentation\cite{kirillov2023segment}. 
Despite the unavailability of SAM's text-based prompting to the public, we adopt alternative strategies such as Grounded-Segment-Anything\cite{ren2024grounded} to achieve comparable outcomes. 
Our approach involves using an open-vocabulary object detector (e.g., Detic\cite{zhou2022detecting} or Grounding DINO\cite{liu2023grounding}) to generate bounding box prompts for SAM. Subsequently, SAM generates masks that enable the precise extraction of pertinent segments from the source imagery. The pipeline is illustrated in Fig. \ref{fig_pip}.

\subsection{Error Handling}
\begin{figure}[tbp]
\includegraphics[width=0.45\textwidth]{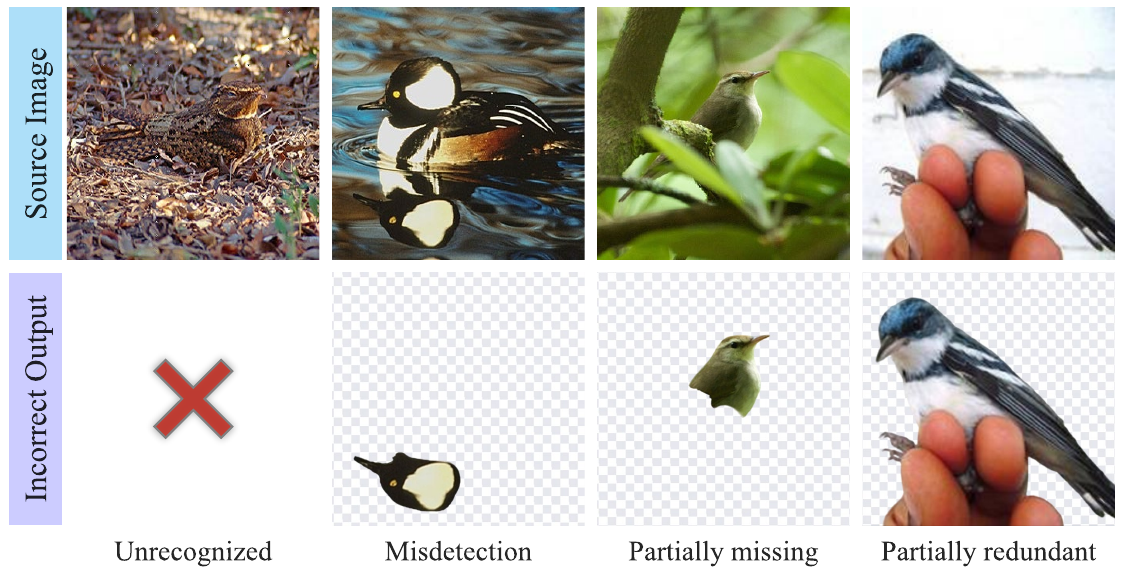}
\centering
\caption{Error handling.} 
\label{fig_err}
\end{figure}

Despite the high accuracy of Grounding-DINO's detection prompts and SAM's segmentation, minor errors (about 1\%-2\%) still occur.
These include unrecognized subjects, incorrect identifications, unwanted background, or incomplete objects, as shown in Fig.\ref{fig_err}. We manually review and correct these issues to ensure the dataset meets high-quality standards, guiding future users in constructing datasets with minimal manual checks.

\subsection{Implementation Details}
To achieve high-quality images while balancing hardware requirements, the pipeline utilizes the ViT-L SAM and Detic\_LI21k\_CLIP\_SwinB models, processed on two NVIDIA 3090s at 0.3s per image. The aspect ratio of images is preserved for seamless dataset replacement, with some images displayed as squares for aesthetic purposes in the paper.

\section{Overview of Processed Datasets}
In this section, we first show examples of foreground datasets aggregated to subjects obtained through the Pipeline preprocessing process described above, before further eliciting information on how it can be used to more conveniently extend and enrich more modalities.

\begin{table}[tbp]
    \centering
    \caption{Source datasets statistics.}
    \resizebox{0.8\columnwidth}{!}{
    \begin{tabular}{c|cc}
    \hline
    \textbf{Dataset} & \textbf{Num. of Classes} & \textbf{Num. of Images} \\ \hline
    CUB 200-2011     & 200                      & 11788                   \\
    Stanford Cars    & 196                      & 16185                   \\
    Aircraft         & 100                      & 10000                   \\
    \hline
    \end{tabular}
    }
\end{table}

\subsection{Source Datasets}
To provide a comprehensive demonstration of the generalisability of the proposed pipeline in different situations, we utiliz a set of the most commonly used benchmark datasets that cover a wide range of fine-grained visual application scenarios. Details of the selected source datasets are given below:

\begin{enumerate}[leftmargin=15pt,itemsep=0pt,topsep=3pt,partopsep=0pt,parsep=1pt]
    \item[{\it 1)}] {\it CUB (Caltech-UCSD Birds-200-2011)\cite{wah2011caltech}:}
    \begin{itemize}[leftmargin=1pt,itemsep=0pt,topsep=0pt,partopsep=0pt,parsep=0pt]
        \item[-] Contains 11,788 images covering 200 bird species.
        \item[-] Renowned for its highly granular categories and superior image quality.
    \end{itemize}

    \item[{\it 2)}] \textit{Stanford-Cars\cite{krause20133d}:}
    \begin{itemize}[leftmargin=1pt,itemsep=0pt,topsep=0pt,partopsep=0pt,parsep=0pt]
        \item[-] Comprises 16,185 images of 196 distinct car categories.
        \item[-] Challenges models with fine-grained recognition tasks, testing their ability to discern subtle differences between car models.
    \end{itemize}

    \item[{\it 2)}] \textit{Aircraft\cite{maji2013fine}:}
    \begin{itemize}[leftmargin=1pt,itemsep=0pt,topsep=0pt,partopsep=0pt,parsep=0pt]
        \item[-] Includes 10,000 images of 100 aircraft models.
        \item[-] Provides a challenging dataset due to diverse aircraft types and viewpoints.
    \end{itemize}
    
\end{enumerate}



    

\subsection{Data Examples}
\begin{figure}[tbp]
\includegraphics[width=0.48\textwidth]{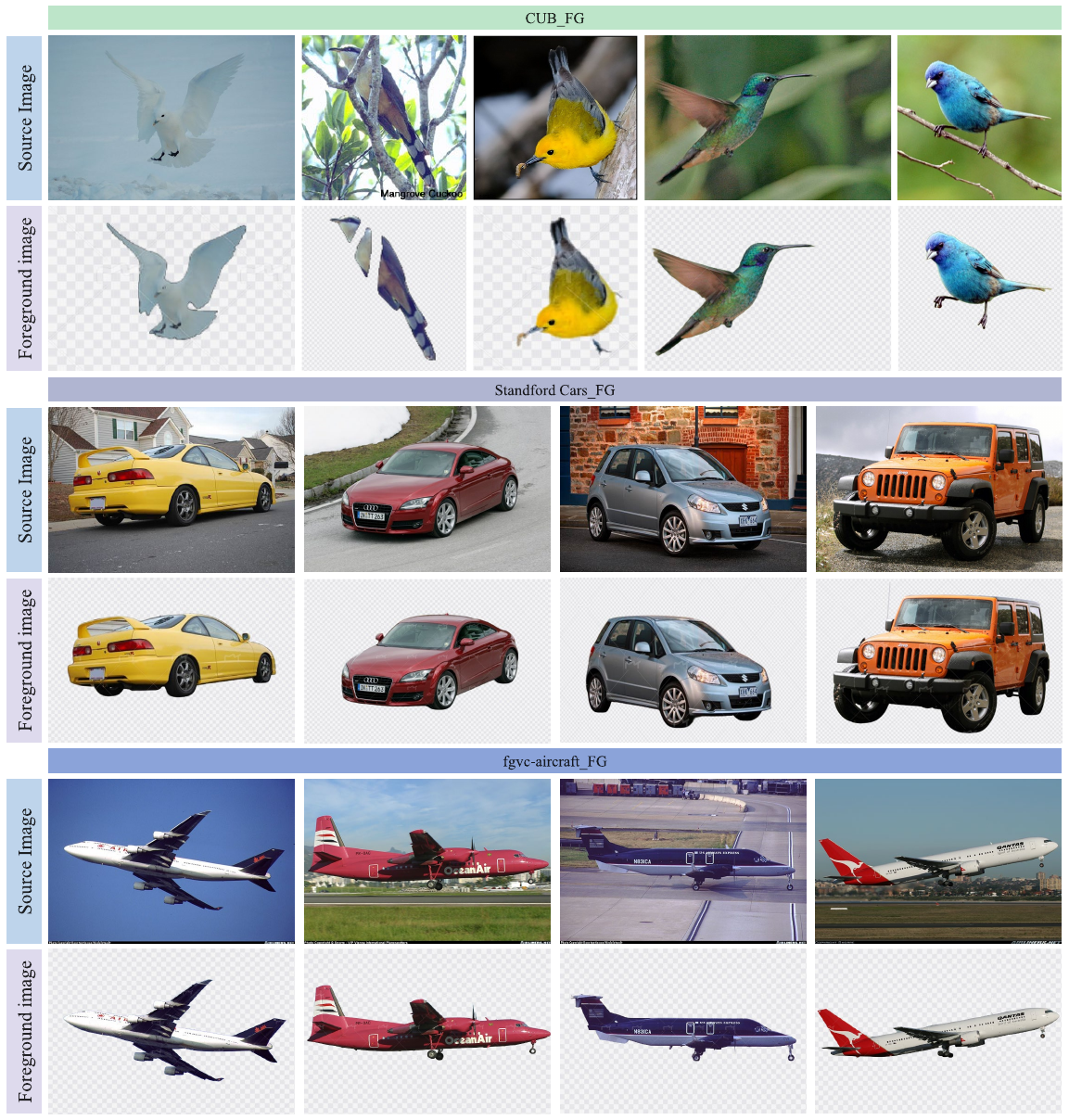}
\centering
\caption{Example of foreground data, including corresponding source images.} 
\label{fig_exm}
\end{figure}

As shown in Fig. \ref{fig_exm}, we showcase partial images from our newly generated foreground-only data for CUB, Standford Cars, and Aircraft, alongside their corresponding source images. 
The resulting images exhibit exceptional quality and sophisticated segmentation, as evidenced by a) the complete removal of background
(e.g., the second image of CUB\_FG, where the branches obstructing the subject are effectively removed); and b) the preservation of all discriminative features of the subject (e.g., the last image of CUB\_FG,in which the tiny but discriminative bird claws are preserved).

\subsection{Modalities Expansion Potential}
Moreover, our newly processed data enables representation of higher quality samples, facilitating the generation of additional modalities to expand tasks and datasets.

\begin{figure}[tbp]
\includegraphics[width=0.42\textwidth]{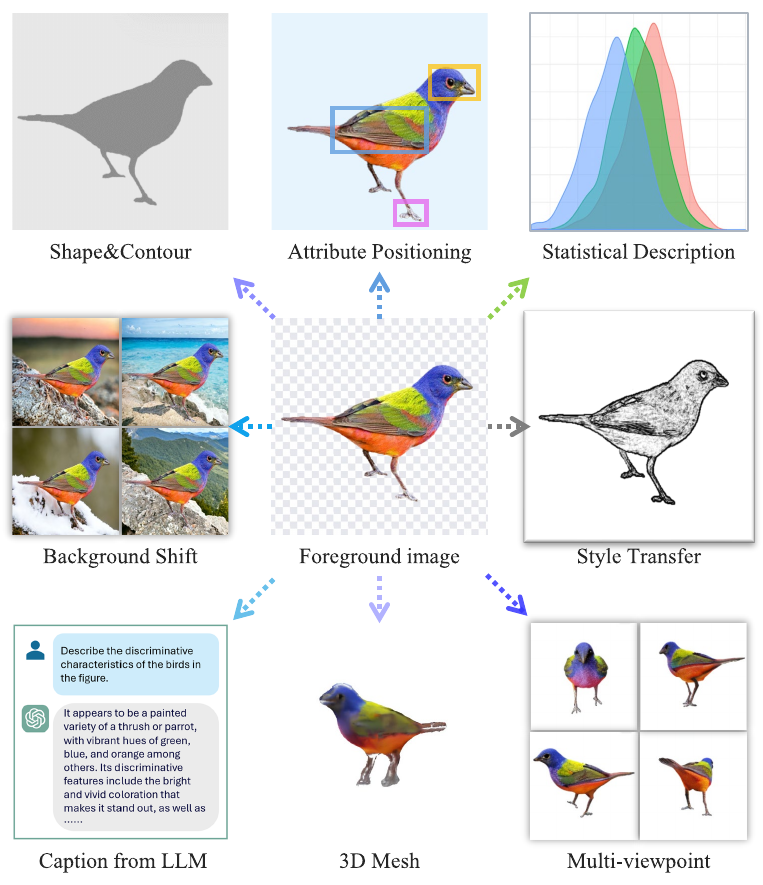}
\centering
\caption{Extending more modalities using foreground images.} 
\label{fig_mm}
\end{figure}

The illustration in Fig. \ref{fig_mm} elucidates the advantages of utilizing foreground-only images, which facilitate the acquisition of:
a) Precise shape and contour information intrinsic to the instance\cite{tang2023weakly}.
b) Enhanced image statistics such as color histograms and SIFT features, tailored to the instance and unaffected by background interference.
c) Improved accuracy in the localization of discriminative attributes.
d) Enhanced input for Large Multimodal Models (e.g., GPT-4\footnote{\url{https://openai.com/index/gpt-4/}}) to produce more precise sample descriptions.
e) 3D mesh generation\cite{wu2024unique3d} through monocular vision-based reconstruction techniques\footnote{\url{https://huggingface.co/spaces/Wuvin/Unique3D}}, allowing for the synthesis of images from various perspectives\cite{gan2024fine}.
f) Effortless background replacement to assess model resilience to background variations.
g) Style transfer of subjects (e.g., watercolor, sketch) to forge new domains, thereby advancing domain adaptation research.

These streamlined approaches underscores the potential of foreground data in advancing various applications.

\section{Experiments}
In this section, we present preliminary experiments on the preprocessing approach, with further explorations reserved for future studies.

\subsection{Feature Distribution}
To assess the impact of background removal on dataset feature distribution, we utilized t-SNE\cite{van2008visualizing} for visualizing high-level features extracted by a pre-trained DINOv2 [ViT-B/16] backbone\cite{oquab2023dinov2}. Note that backbone is not fine-tuned using the corresponding data, thus ensuring the generality of the features. 

\begin{figure}[tbp]
\includegraphics[width=0.45\textwidth]{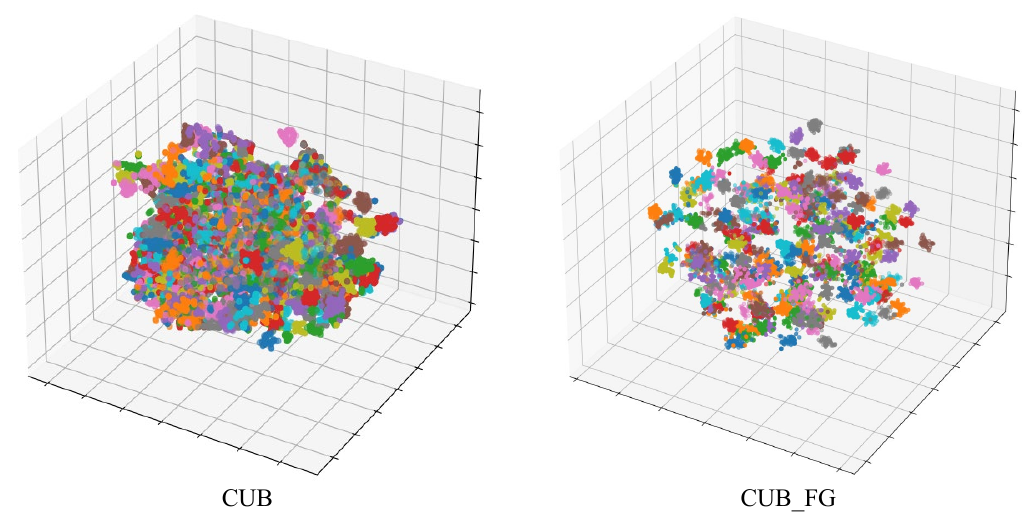}
\centering
\caption{The t-SNE visualisation of high-level features for the CUB and CUB\_FG datasets.} \label{fig_vis}
\end{figure}

Fig. \ref{fig_vis} contrasts the CUB\_FG dataset with the original CUB, highlighting a tighter intra-subclass clustering and an expanded inter-subclass separation post-background removal. This suggests that CUB\_FG represents an enhanced dataset, facilitating the investigation of model capabilities in extracting discriminative features without background noise interference.


\subsection{Cross-validation Experiments}
To validate the improvements of the new dataset over the original and to demonstrate the previously discussed influence of background elements on model performance in fine-grained classification tasks, we conducted a comparative analysis. Employing a classic fine-grained classification task as a reference, we tested common backbone architectures, ViT-Base/16\cite{dosovitskiy2020image}, ResNet-50\cite{he2016deep}, Swin Transformer v2-Base\cite{liu2022swin}, and ConvNeXt-Base\cite{liu2022convnet}, using their standard configurations and pre-trained weights, devoid of additional tricks or modules. As an example, we fine-tuned these models on the CUB\_FG training set and evaluated their performance on the original CUB test set.

\begin{figure}[tbp]
\includegraphics[width=0.5\textwidth]{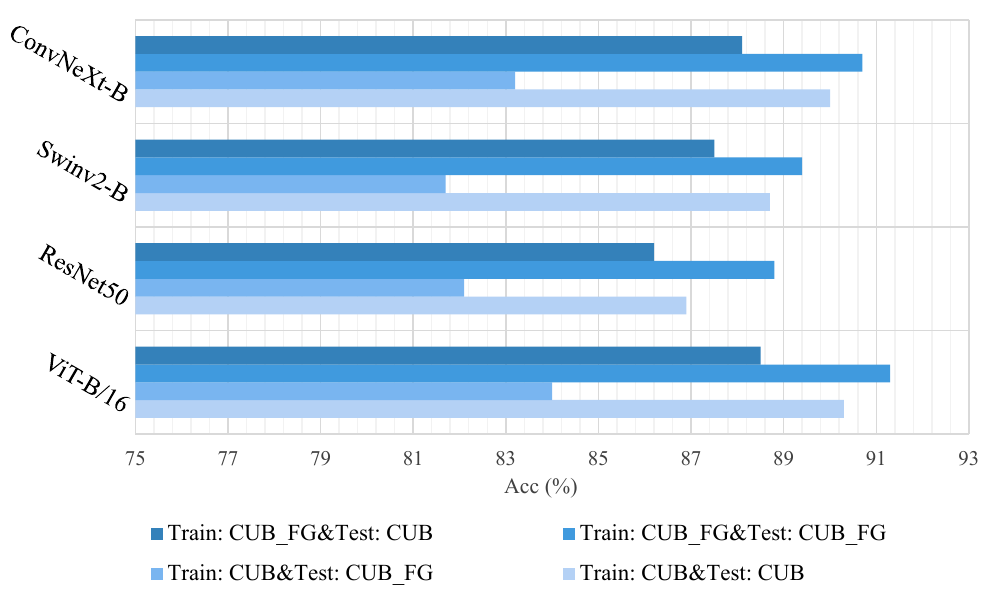}
\centering
\caption{Classification Acc. of cross-validation experiments on CUB and CUB\_FG datasets.} \label{fig_crs}
\end{figure}

The cross-validation outcomes are detailed in Table \ref{cross-ex}, with corresponding bar charts for the CUB and CUB\_FG datasets depicted in Fig.\ref{fig_crs} for clarity. Key findings include:
\begin{itemize}[leftmargin=15pt]
\item Across the four backbones and three datasets,  a substantial decrease in classification accuracy (over 6\% on average) when models trained on the source data are evaluated on the foreground data, validating our hypothesis regarding the detrimental impact of background noise on model focus.
\item Models trained on the foreground datasets consistently outperformed those trained on the source datasets, indicating the higher quality of the foreground datasets for training and supporting the utility of our preprocessing pipeline in boosting model performance.
\item A modest performance drop (within 2.5\% on average in classification accuracy) when foreground-trained models are tested on the source datasets, attributed to data format discrepancies between training and testing phases.
\item The performance improvement of models trained on the foreground datasets using the Transformer-based ViT architecture was notably pronounced. Benefiting from its robust integration of global information, the removal of background noise further enhanced its performance.
\end{itemize}

\begin{table}[!t]
\caption{Cross-validation experiments on each dataset.}
\label{cross-ex}
\resizebox{\columnwidth}{!}{%
\begin{tabular}{cc|cccc}
\hline
\multicolumn{2}{c|}{Dataset}                     & \multicolumn{4}{c}{Backbone Arch. (Acc. \%)}           \\ \hline
\multicolumn{1}{c|}{Train}        & Test         & ViT-B/16 & ResNet-50 & Swinv2-B & ConvNeXt-B \\ \hline
\multicolumn{1}{c|}{CUB}          & CUB          & 90.3     & 86.9      & 88.7     & 90.0       \\
\multicolumn{1}{c|}{CUB}          & CUB\_FG      & 84.0     & 82.1      & 81.7     & 83.2       \\
\multicolumn{1}{c|}{CUB\_FG}      & CUB\_FG      & 91.3     & 88.8      & 89.4     & 90.7       \\
\multicolumn{1}{c|}{CUB\_FG}      & CUB          & 88.5     & 86.2      & 87.5     & 88.1       \\ \hline
\multicolumn{1}{c|}{Cars}         & Cars         & 94.4     & 94.0      & 91.8     & 93.6       \\
\multicolumn{1}{c|}{Cars}         & Cars\_FG     & 88.6     & 88.5      & 86.4     & 89.0       \\
\multicolumn{1}{c|}{Cars\_FG}     & Cars\_FG     & 95.3     & 95.1      & 94.1     & 94.6       \\
\multicolumn{1}{c|}{Cars\_FG}     & Cars         & 93.2     & 93.7      & 91.5     & 92.9       \\ \hline
\multicolumn{1}{c|}{Aircraft}     & Aircraft     & 92.8     & 92.4      & 90.4     & 92.6       \\
\multicolumn{1}{c|}{Aircraft}     & Aircraft\_FG & 87.3     & 88.5      & 87.0     & 87.6       \\
\multicolumn{1}{c|}{Aircraft\_FG} & Aircraft\_FG & 93.8     & 93.1      & 91.6     & 93.1       \\
\multicolumn{1}{c|}{Aircraft\_FG} & Aircraft     & 90.2     & 89.7      & 90.0     & 90.3       \\ \hline
\end{tabular}%
}
\end{table}

\section{Conclusion}
In this paper, we introduced an engineered pipeline to aid in the fine-grained analysis of how background influences model performance and to explore the ability of models to identify truly discriminative features focused on the foreground. Leveraging the capabilities of the Detic\cite{zhou2022detecting} and SAM\cite{kirillov2023segment}, we constructed data that isolate the foreground subjects by removing the background from the original images. 
Initial explorations with diverse backbone models on these datasets have been conducted, providing a solid foundation for further research on adaptive feature extraction methods that are insensitive to background changes.

Moving forward, we plan to expand these datasets across modalities to bolster foundational research and investigate adaptive feature extraction methods robust to background influences.

\vfill
\pagebreak

\bibliographystyle{IEEEbib}
\bibliography{strings,ref}

\begin{thebibliography}{10}

\bibitem{xu2023mmcosine}
Ruize Xu, Ruoxuan Feng, Shi-Xiong Zhang, and Di~Hu,
\newblock ``Mmcosine: Multi-modal cosine loss towards balanced audio-visual fine-grained learning,''
\newblock in {\em ICASSP 2023-2023 IEEE International Conference on Acoustics, Speech and Signal Processing (ICASSP)}. IEEE, 2023, pp. 1--5.

\bibitem{he2022transfg}
Ju~He, Jie-Neng Chen, Shuai Liu, Adam Kortylewski, Cheng Yang, Yutong Bai, and Changhu Wang,
\newblock ``Transfg: A transformer architecture for fine-grained recognition,''
\newblock in {\em Proceedings of the AAAI Conference on Artificial Intelligence}, 2022, vol.~36, pp. 852--860.

\bibitem{zha2023boosting}
Zican Zha, Hao Tang, Yunlian Sun, and Jinhui Tang,
\newblock ``Boosting few-shot fine-grained recognition with background suppression and foreground alignment,''
\newblock {\em IEEE Transactions on Circuits and Systems for Video Technology}, vol. 33, no. 8, pp. 3947--3961, 2023.

\bibitem{selvaraju2017grad}
Ramprasaath~R Selvaraju and et~al.,
\newblock ``Grad-cam: Visual explanations from deep networks via gradient-based localization,''
\newblock in {\em ICCV}, 2017, pp. 618--626.

\bibitem{kirillov2023segment}
Alexander Kirillov, Eric Mintun, Nikhila Ravi, Hanzi Mao, Chloe Rolland, Laura Gustafson, Tete Xiao, Spencer Whitehead, Alexander~C Berg, Wan-Yen Lo, et~al.,
\newblock ``Segment anything,''
\newblock in {\em Proceedings of the IEEE/CVF International Conference on Computer Vision}, 2023, pp. 4015--4026.

\bibitem{zhou2022detecting}
Xingyi Zhou, Rohit Girdhar, Armand Joulin, Philipp Kr{\"a}henb{\"u}hl, and Ishan Misra,
\newblock ``Detecting twenty-thousand classes using image-level supervision,''
\newblock in {\em European Conference on Computer Vision}. Springer, 2022, pp. 350--368.

\bibitem{ren2024grounded}
Tianhe Ren, Shilong Liu, Ailing Zeng, Jing Lin, Kunchang Li, He~Cao, Jiayu Chen, Xinyu Huang, Yukang Chen, Feng Yan, Zhaoyang Zeng, Hao Zhang, Feng Li, Jie Yang, Hongyang Li, Qing Jiang, and Lei Zhang,
\newblock ``Grounded sam: Assembling open-world models for diverse visual tasks,'' 2024.

\bibitem{liu2023grounding}
Shilong Liu, Zhaoyang Zeng, Tianhe Ren, Feng Li, Hao Zhang, Jie Yang, Chunyuan Li, Jianwei Yang, Hang Su, Jun Zhu, et~al.,
\newblock ``Grounding dino: Marrying dino with grounded pre-training for open-set object detection,''
\newblock {\em arXiv preprint arXiv:2303.05499}, 2023.

\bibitem{wah2011caltech}
Catherine Wah and et~al.,
\newblock ``The caltech-ucsd birds-200-2011 dataset,''
\newblock 2011, California Institute of Technology.

\bibitem{krause20133d}
Jonathan Krause, Michael Stark, Jia Deng, and Li~Fei-Fei,
\newblock ``3d object representations for fine-grained categorization,''
\newblock in {\em ICCV Workshop}, 2013, pp. 554--561.

\bibitem{maji2013fine}
Subhransu Maji and et~al.,
\newblock ``Fine-grained visual classification of aircraft,''
\newblock 2013.

\bibitem{tang2023weakly}
Zhenchao Tang, Hualin Yang, and Calvin Yu-Chian Chen,
\newblock ``Weakly supervised posture mining for fine-grained classification,''
\newblock in {\em Proceedings of the IEEE/CVF Conference on Computer Vision and Pattern Recognition}, 2023, pp. 23735--23744.

\bibitem{wu2024unique3d}
Kailu Wu, Fangfu Liu, Zhihan Cai, Runjie Yan, Hanyang Wang, Yating Hu, Yueqi Duan, and Kaisheng Ma,
\newblock ``Unique3d: High-quality and efficient 3d mesh generation from a single image,''
\newblock {\em arXiv preprint arXiv:2405.20343}, 2024.

\bibitem{gan2024fine}
Qijun Gan, Wentong Li, Jinwei Ren, and Jianke Zhu,
\newblock ``Fine-grained multi-view hand reconstruction using inverse rendering,''
\newblock in {\em Proceedings of the AAAI Conference on Artificial Intelligence}, 2024, vol.~38, pp. 1779--1787.

\bibitem{van2008visualizing}
Laurens Van~der Maaten and Geoffrey Hinton,
\newblock ``Visualizing data using t-sne.,''
\newblock {\em Journal of machine learning research}, vol. 9, no. 11, 2008.

\bibitem{oquab2023dinov2}
Maxime Oquab, Timoth{\'e}e Darcet, Th{\'e}o Moutakanni, Huy~V Vo, Marc Szafraniec, Vasil Khalidov, Pierre Fernandez, Daniel HAZIZA, Francisco Massa, Alaaeldin El-Nouby, et~al.,
\newblock ``Dinov2: Learning robust visual features without supervision,''
\newblock {\em Transactions on Machine Learning Research}, 2023.

\bibitem{dosovitskiy2020image}
Alexey Dosovitskiy, Lucas Beyer, Alexander Kolesnikov, Dirk Weissenborn, Xiaohua Zhai, Thomas Unterthiner, Mostafa Dehghani, Matthias Minderer, Georg Heigold, Sylvain Gelly, et~al.,
\newblock ``An image is worth 16x16 words: Transformers for image recognition at scale,''
\newblock in {\em International Conference on Learning Representations}, 2020.

\bibitem{he2016deep}
Kaiming He, Xiangyu Zhang, Shaoqing Ren, and Jian Sun,
\newblock ``Deep residual learning for image recognition,''
\newblock in {\em Proceedings of the IEEE conference on computer vision and pattern recognition}, 2016, pp. 770--778.

\bibitem{liu2022swin}
Ze~Liu, Han Hu, Yutong Lin, Zhuliang Yao, Zhenda Xie, Yixuan Wei, Jia Ning, Yue Cao, Zheng Zhang, Li~Dong, et~al.,
\newblock ``Swin transformer v2: Scaling up capacity and resolution,''
\newblock in {\em Proceedings of the IEEE/CVF conference on computer vision and pattern recognition}, 2022, pp. 12009--12019.

\bibitem{liu2022convnet}
Zhuang Liu, Hanzi Mao, Chao-Yuan Wu, Christoph Feichtenhofer, Trevor Darrell, and Saining Xie,
\newblock ``A convnet for the 2020s,''
\newblock in {\em Proceedings of the IEEE/CVF conference on computer vision and pattern recognition}, 2022, pp. 11976--11986.

\end{thebibliography}

\end{document}